\documentclass[pdflatex,sn-vancouver-num]{sn-jnl}%
\usepackage{graphicx}%
\usepackage{multirow}%
\usepackage{amsmath,amssymb,amsfonts}%
\usepackage{amsthm}%
\usepackage{mathrsfs}%
\usepackage[title]{appendix}%
\usepackage{xcolor}%
\usepackage{textcomp}%
\usepackage{manyfoot}%
\usepackage{booktabs}%
\usepackage{algorithm}%
\usepackage{algorithmicx}%
\usepackage{algpseudocode}%
\usepackage{listings}%

\theoremstyle{thmstyleone}%

\theoremstyle{thmstyletwo}%

\theoremstyle{thmstylethree}%

\begin{document}

\title[Short Title]{Enhancing Mental Health Classification with Layer-Attentive Residuals and Contrastive Feature Learning}
\author*[1]{Menna Elgabry}
\email{mennayasser891@gmail.com}

\author[1]{Ali Hamdi}
\email{ahamdi@msa.edu.eg}

\author[2]{Khaled Shaban}
\email{khaled.shaban@qu.edu.qa}

\affil*[1]{\orgdiv{Department of Computer Science},
\orgname{Modern Sciences and Arts University},
\orgaddress{\city{Cairo}, \country{Egypt}}}

\affil[2]{\orgdiv{Computer Science and Engineering Department},
\orgname{College of Engineering, Qatar University},
\orgaddress{\city{Doha}, \country{Qatar}}}

\abstract{The classification of mental health is challenging for a variety of reasons. For one, there is overlap between the mental health issues. In addition, the signs of mental health issues depend on the context of the situation, making classification difficult. Although fine-tuning transformers has improved the performance for mental health classification, standard cross-entropy training tends to create entangled feature spaces and fails to utilize all the information the transformers contain. We present a new framework that focuses on representations to improve mental health classification. This is done using two methods. First, \textbf{layer-attentive residual aggregation} which works on residual connections to to weigh and fuse representations from all transformer layers while maintaining high-level semantics. Second, \textbf{supervised contrastive feature learning} uses temperature-scaled supervised contrastive learning with progressive weighting to increase the geometric margin between confusable mental health problems and decrease class overlap by restructuring the feature space. With a score of \textbf{74.36\%}, the proposed method is the best performing on the SWMH benchmark and outperforms models that are domain-specialized, such as \textit{MentalBERT} and \textit{MentalRoBERTa} by margins of (3.25\% - 2.2\%) and 2.41 recall points over the highest achieving model. These findings show that domain-adaptive pretraining for mental health text classification can be surpassed by carefully designed representation geometry and layer-aware residual integration, which also provide enhanced interpretability through learnt layer importance.
}


\maketitle

\section{Introduction}
Over one billion people worldwide suffer from mental health issues, which is a global crisis and a major contributor to the burden of disability. People are increasingly using digital narratives to convey psychological suffering, which has led to new opportunities for mental health surveillance due to the development of social media platforms \cite{doi:10.1177/1178222618792860, first}. Mental health text categorization is becoming a crucial natural language processing (NLP) task with significant practical implications for early detection, intervention, and public health monitoring due to the convergence of need and data availability \cite{ji-etal-2022-mentalbert}.

Mental health classification is still particularly difficult, despite significant advancements in transformer-based methods. The task functions at the nexus of ethical sensitivity, psychological complexity, and linguistic intricacy. In contrast to general text classification, mental health disorders frequently show up as context-dependent symptomatology, comorbidity patterns, and subtle expressions 
\cite{inproceedings}. For example, standard fine-tuned transformers frequently fail to capture the difference between generalized anxiety ("I constantly worry about everything") and depression ("I feel empty inside"), which requires sensitivity to both semantic content and pragmatic framing \cite{benton2017multitasklearningmentalhealth}.

Current methods continue to have two basic drawbacks. Initially, \textbf{representation entanglement}: Semantically comparable but clinically dissimilar diseases typically occupy overlapping regions in feature space in transformer models that have been fine-tuned using standard cross-entropy loss. Unstable decision boundaries result from this entanglement, especially for disorders like anxiety and depression that have similar symptom characteristics \cite{shing-etal-2018-expert}. Secondly, \textbf{inefficient layer utilization}: Although transformer architectures generate rich hierarchical representations across layers (with higher layers encoding semantic relationships and lower layers capturing syntactic patterns) \cite{jawahar-etal-2019-bert}, the majority of mental health classification approaches either only use the final layer or use basic aggregation techniques that do not fully utilize this hierarchical richness \cite{zhuang-etal-2021-robustly}.

In this work, a novel system (Layer-Attentive Residuals with Contrastive learning) is introduced. It uses two complimentary innovations to address these shortcomings. In order to enforce intra-class compactness and inter-class separation, we first apply a \textbf{temperature-scaled supervised contrastive loss} that explicitly shapes the representation space. In contrast to conventional contrastive methods, our technique uses progressive weighting and adaptive temperature scaling to account for the special features of mental health text, where certain disorders are inherently closer in symptom space than others. Second, given that different transformer layers encode syntactic and semantic information at varying depths \cite{rogers-etal-2020-primer} we suggest a \textbf{learnable layer-attention mechanism} that dynamically assigns weights to representations from every transformer layer according to how important they are for classifying mental health. By acting as a kind of residual connection throughout the transformer hierarchy, this attention mechanism enables the model to downweight less informative layers and selectively highlight layers that capture clinically significant data.

We implement our method on the RoBERTa model \cite{zhuang-etal-2021-robustly} and assess it on the SWMH benchmark \cite{ji2021suicidal}, which is an extensive dataset of four mental health problems and a control class. Our approach surpasses numerous strong baselines, including MentalBERT \cite{ji-etal-2022-mentalbert}, and achieves \textbf{74.36\% F1-score}, which is also an improvement over traditional RoBERTa fine-tuning baseline.

Our approach offers two types of interpretability that are useful for clinical applications, in addition to quantitative benefits. The learnt layer attention weights provide transparency into model decision-making by revealing which linguistic factors (lexical, syntactic, and semantic) influence predictions. In the meantime, the contrastive projections provide qualitative confirmation of representation quality by producing visualizable embeddings where clinically linked circumstances cluster correctly.

\textbf{Our contributions are threefold:}
\begin{enumerate} 

\item In order to address the complex interactions between psychological states, a novel contrastive learning formulation designed especially for mental health text categorization incorporates progressive weighting and adaptive temperature scaling.
    
    \item A dynamic layer-attention technique that improves performance and interpretability by learning to weight transformer layer representations according to their clinical relevance, allowing language features to influence classification judgments.
    
    \item Ablation tests reveal that each of these components considerably improves overall performance, providing thorough empirical evidence that these components are complimentary. 
\end{enumerate}

This study brings us closer to trustworthy, interpretable tools for digital mental health assessment—tools that not only accurately categorize but also do so in ways that doctors can comprehend and trust—by connecting representation learning advancements with the particular requirements of mental health text analysis.

\section{Related Work}

\subsection{Mental Health Text Classification}

Over the past ten years, computational methods for evaluating mental health from text have greatly changed. \citet{doi:10.1177/1178222618792860} pioneered the use of social media data for suicide risk assessment through meticulously built psycholinguistic characteristics, while earlier work relied on lexicons and manufactured linguistic features. Although this field of study demonstrated the feasibility of automated mental health screening, it was constrained by domain specialization and feature engineering overhead.

Significant advancements were made with the introduction of neural architectures. \citet{benton2017multitasklearningmentalhealth} showed that shared representation learning could enhance performance in multitask learning across various mental health issues. Later research investigated different brain architectures, concentrating on the identification of anxiety and depression as high-prevalence states \cite{shing-etal-2018-expert}. However, these methods frequently addressed each illness separately, ignoring the intricate comorbidity patterns that define mental health in the actual world.

Transformer-based models represented a major advancement. \citet{ji-etal-2022-mentalbert} shown that domain-adaptive pretraining produces significant benefits by introducing MentalBERT, a RoBERTa model continuously pretrained on mental health forums. At the same time, \citet{inproceedings} investigated domain adaptation methods for depression identification, emphasizing the difficulties in transferring knowledge between various text genres (e.g., social media vs. clinical notes). Despite achieving state-of-the-art performance, these transformer techniques mostly adopted the conventional fine-tuning paradigm without tackling the underlying issues with representation structure for mental health disorders.

\subsection{Contrastive Learning in NLP}

Contrastive learning is a powerful approach for learning well-structured representations. When samples from the same class are purposefully drawn together while samples from different classes are pushed apart, the supervised contrastive loss demonstrates that representations with higher generalization properties are created \cite{khosla2021supervisedcontrastivelearning}. Separable decision boundaries have made this approach particularly effective.

For a number of NLP issues, contrastive approaches have been updated. As shown by \citet{gao-etal-2021-simcse}, simple contrastive learning objectives might produce high-quality sentence embeddings without supervised labeling. It has been shown that the temperature parameter in contrastive learning, which controls the distribution's concentration, is crucial for performance \cite{khosla2021supervisedcontrastivelearning}, with various tasks requiring various temperature calibrations.

There is a noticeable scarcity of work in this discipline that addresses the particular challenges of mental health text, where disorders exhibit complex, overlapping symptomatology rather than clear category boundaries. Our work extends supervised contrastive learning to the field of mental health classification using adaptive temperature scaling and progressive weighting methods tailored to its unique characteristics.

\subsection{Transformer Layer Analysis and Utilization}

Different layers collect different language information, according to analysis of transformer architectures. \citet{jawahar-etal-2019-bert} methodically shown that whereas higher levels record semantic linkages, lower layers mostly encode syntactic information. This hierarchical structure implies that rather than depending only on the output of the last layer, good performance on complicated tasks may involve combining information across numerous layers.

Additionally, \citet{tenney-etal-2019-bert} showed that BERT's layers roughly rebuild the conventional NLP pipeline, with semantic roles emerging in later layers, parsing in middle layers, and part-of-speech tagging emerging in early stages. Because different tasks may benefit from emphasizing different levels, this finding has significant implications for task-specific model design.

Practical strategies for utilizing multi-layer information are still somewhat basic in spite of these discoveries. Typical methods include employing all layers with equal weighting or concatenating the final four layers \cite{zhuang-etal-2021-robustly}. However, more advanced techniques, such as learnable layer attention, have not been consistently used in the classification of texts related to mental health, where various illnesses may depend on distinct language signals recorded at various levels of abstraction.

\section{Methodology}

\subsection{Problem Formulation}

Classifying texts related to mental health has special difficulties that go beyond typical multi-class classification. Learning representations whose geometric form reflects the clinical correlations between diseases is of greater importance to us than just label prediction. Formally, a dataset of $N$ text samples is represented as $\mathcal{D} = (\mathbf{x}_i, y_i)\}_{i=1}^N$, where $\mathbf{x}_i$ is a text sequence and $y_i \in\,1, \dots, K\}$ is the matching mental health condition label from $K$ potential classes. After tokenization, each text $\mathbf{x}_i = [x_{i,1}, \dots, x_{i,T}]$ consists of $T$ tokens. We have two objectives: (1) to discover a function $f: \mathcal{X} \rightarrow \mathbb{R}^K$ that converts text inputs into precise probability scores across classes, and (2) to guarantee that the internal representations generated by $f$ provide a semantically meaningful space in which clinically dissimilar conditions are farther distant from those with comparable symptomatology.

We assess our method on the SWMH benchmark \cite{ji2021suicidal}, which is an extensive dataset of  54,412 user posts sourced from mental health-related subreddits that has four mental health problems together with a control class. Training, validation, and test sets included roughly 63.9\%, 16.0\%, and 20.0\% of the total data, respectively. This translates to approximately 10.9k examples for testing, 8.71k examples for validation, and 34.8k examples for training. 

\subsection{Base Architecture: RoBERTa-Large}

We use RoBERTa-large \cite{zhuang-etal-2021-robustly} as our basic language model because of its demonstrated efficacy on social media writing and its ability to comprehend complex semantic information. The transformer model RoBERTa-large has 16 attention heads, $L = 24$ layers, and a hidden dimension of $d = 1024$. At each of its layers, the model produces a series of contextualized representations for an input sequence $\mathbf{x}_i$:

\[
\mathbf{H}_i^l = \text{RoBERTa}_l(\mathbf{x}_i) \in \mathbb{R}^{T \times d}, \quad l=1,\dots,L
\]

The matrix of hidden states for all $T$ tokens at transformer layer $l$ is represented here by $\mathbf{H}_i^l$. We employ the representation of the unique `[CLS]` token as a synopsis of the complete input sequence, in accordance with standard procedure in sentence classification with transformers. This is extracted from each layer: $\mathbf{h}_i^l = \mathbf{H}_i^l[0] \in \mathbb{R}^d$. This provides us with a rich collection of 24 layer-specific `[CLS]` vectors, each of which captures linguistic data at a distinct abstraction level.

\subsection{Layer-Attentive Residual Connections}

Standard fine-tuning's primary drawback is that it ignores the rich, hierarchical information dispersed throughout all transformer stages in favor of relying just on the output of the last layer. Surface and syntactic information are captured by earlier levels, semantic links are captured by middle layers, and task-specific signals are integrated by later layers. This hierarchy is important when it comes to mental health texts. While some illnesses may depend more on deeper semantic content, others may rely more on syntactic patterns of discomfort.

\subsubsection{Multi-Layer Representation Aggregation}

To take advantage of representations from every transformer layer, we present a learnable layer-attention technique. Let $\mathbf{h}_i^{l}$ represent the $l$-th layer's \texttt{[CLS]} representation. A collection of learnable scalar parameters $\mathbf{w} = [w_1, \dots, w_{L+1}] \in \mathbb{R}^{L+1}$ is defined, with an extra parameter $w_{L+1}$ acting as an extra layer and corresponding to the final pooled output. A softmax is used to calculate the normalized importance weights:

\[
\alpha_l = \frac{\exp(w_l)}{\sum_{j=1}^{L+1} \exp(w_j)}, \quad l = 1, \dots, L+1
\]

Each layer's relative contribution is determined by the weights $\alpha_l$, which are learned simultaneously with the model parameters and initialized uniformly. This process, which is inspired by the information bottleneck principle, pushes the model to suppress less informative characteristics and highlight task-relevant representations. Consequently, layers providing stronger semantic signals obtain greater weights, whilst less important shallow representations are attenuated.

\subsubsection{Weighted Feature Fusion}

Next, using the learned attention scores as a guide, we calculate a fused representation $\mathbf{z}_i^{\text{fused}}$ as a weighted sum of all layer representations:

\[
\mathbf{z}_i^{\text{fused}} = \sum_{l=1}^{L+1} \alpha_l \cdot \mathbf{h}_i^l \in \mathbb{R}^d
\]

Numerical stability is ensured by this convex combination. Because it is task-specific and dynamic, this formulation has great power. When differentiating between sadness and anxiety, for instance, the model may learn to give middle layers (e.g., 12–18) larger weights ($\alpha_l$) since these conditions are frequently distinguished by minor semantic subtleties best captured at those intermediate levels of abstraction.

\subsubsection{Residual Enhancement}

We include a residual connection to guarantee good training and maintain the strong signal from the last, most sophisticated layer. The final representation $\mathbf{z}_i$ that can be used is:

\[
\mathbf{z}_i = \mathbf{z}_i^{\text{fused}} + \text{LayerNorm}(\mathbf{h}_i^{L+1})
\]

In this case, $\text{LayerNorm}$ indicates layer normalization \cite{ba2016layernormalization}, and $\mathbf{h}_i^{L+1}$ is the last layer's `[CLS]` token (or pooled output). In addition to ensuring that the fused representation preserves the high-level, task-adapted properties that the final transformer layer has already learnt, this residual connection mitigates any vanishing gradient problems by guaranteeing a direct gradient path to the final layer.

\subsection{Temperature-Scaled Supervised Contrastive Learning}

Predicting the right label is the only goal of standard cross-entropy loss. The internal geometry of the model's representation space is not specifically shaped by it. Decision boundaries may become unstable as a result of "entangled" representations, when distinct but related symptoms (such as anxiety and depression) are not clearly distinguished. We use a supervised contrastive learning objective that specifically combines representations from the same class and separates those from different classes in order to address this.

\subsubsection{Contrastive Projection Head}

First, we employ a simple, trainable neural network head to project the fused representation $\mathbf{z}_i$ onto a specific contrastive space:

\[
\mathbf{c}_i = \mathbf{W}_2 \cdot \text{GELU}(\mathbf{W}_1 \cdot \mathbf{z}_i + \mathbf{b}_1) + \mathbf{b}_2 \in \mathbb{R}^{d_c}
\]

Here, $\mathbf{W}_1 \in \mathbb{R}^{512 \times d}$, $\mathbf{b}_1 \in \mathbb{R}^{512}$, $\mathbf{W}_2 \in \mathbb{R}^{d_c \times 512}$, $\mathbf{b}_2 \in \mathbb{R}^{d_c}$, with $d_c = 256$. For non-linearity, the GELU activation function \cite{hendrycks2023gaussianerrorlinearunits} is employed. In contrast to the features utilized for the final classification, this projection head enables the model to learn features that are specifically tailored for separation and clustering.

\subsubsection{Adaptive Temperature Scaling}

We use a supervised contrastive loss (SupCon). For a batch $\mathcal{B}$ of size $B$, the loss for an anchor sample $i$ is as shown in figure 1:
\begin{figure*}[t]
\centering
\[
\mathcal{L}_{\text{con}} = \frac{1}{|B|} \sum_{i \in B}
- \frac{1}{|P(i)|} \sum_{p \in P(i)}
\log
\frac{
\exp(\text{sim}(\mathbf{c}_i, \mathbf{c}_p)/\tau_i)
}{
\sum_{a \in B \setminus \{i\}}
\exp(\text{sim}(\mathbf{c}_i, \mathbf{c}_a)/\tau_i)
}
\]
\caption{Supervised contrastive loss with adaptive temperature scaling.}
\label{fig:supcon}
\end{figure*}

Where:
\begin{itemize}
    \item $P(i) = \{p \in \mathcal{B} : y_p = y_i, p \neq i\}$ is the set of indices of all other samples in the batch with the same label as $i$ (the "positives").
    \item $\text{sim}(\mathbf{u}, \mathbf{v}) = \frac{\mathbf{u}^\top \mathbf{v}}{\|\mathbf{u}\|\|\mathbf{v}\|}$ is the cosine similarity.
    \item $\tau_i > 0$ is a \textbf{key innovation}: an adaptive temperature parameter.
\end{itemize}

The "sharpness" of the similarity distribution is determined by the temperature $\tau$. A low $\tau$ makes separation more difficult by amplifying discrepancies. Importantly, not every pair of mental health conditions is equally simple to distinguish. Consequently, we create $\tau$ sample-specific and adaptive:

\[
\tau_i = \tau_{\text{base}} \cdot (1 + \beta \cdot \sigma(\mathbf{z}_i))
\]

with $\tau_{\text{base}} = 0.05$, $\beta = 0.1$, and $\sigma(\cdot)$, a learnt scalar function (a little neural network) that calculates the intrinsic "difficulty" or ambiguity of the input representation $\mathbf{z}_i$. This enables the model to apply a stronger pull/push (lower $\tau$) for situations that are clear-cut and a gentler one (higher $\tau$) for examples that are intrinsically unclear or similar-condition.  Our adaptive temperature scaling can be interpreted from an information-theoretic perspective related to Fisher information. The temperature $\tau$ inversely influences the sharpness of the similarity distribution in the contrastive loss. Conditions with high Fisher information (easier to distinguish) benefit from a lower $\tau$, creating a sharper, more confident separation. Conditions with low Fisher information (highly confusable, like depression and anxiety) require a higher $\tau$ to prevent the loss from becoming too stringent and unstable, which could lead to poor optimization. Our learned $\tau_i$ function approximates this adaptive calibration.

\subsubsection{Progressive Weighting Schedule}

It can be disruptive to start training with the contrastive loss. Therefore, we employ a straightforward linear timetable to progressively enhance its influence:

\[
\lambda(t) = \lambda_{\text{max}} \cdot \min\left(1, \frac{t}{T_{\text{ramp}}}\right)
\]

The current training epoch is denoted by $t$, the ramp-up period is denoted by $T_{\text{ramp}} = 5$, and the maximum weight is denoted by $\lambda_{\text{max}} = 0.15$. This enables the model to use cross-entropy to build a fundamental classification foundation before using the contrastive signal to further refine the representation geometry.

\subsection{Classification Head and Joint Optimization}

\subsubsection{Enhanced Classification Head}

For the final classification decision, we process the fused representation $\mathbf{z}_i$ (not the contrastive projection $\mathbf{c}_i$) through a separate classification head:

\[
\mathbf{o}_i = \mathbf{W}_4 \cdot \text{Dropout}(\text{GELU}(\mathbf{W}_3 \cdot \mathbf{z}_i + \mathbf{b}_3)) + \mathbf{b}_4 \in \mathbb{R}^K
\]

The matrices are $\mathbf{W}_3 \in \mathbb{R}^{512 \times d}$, $\mathbf{b}_3 \in \mathbb{R}^{512}$, $\mathbf{W}_4 \in \mathbb{R}^{K \times 512}$, $\mathbf{b}_4 \in \mathbb{R}^K$. A dropout rate of 0.4 is applied for regularization. The output $\mathbf{o}_i$ are the logits for the $K$ mental health conditions.

\subsubsection{Joint Loss Function}

The weighted sum of our contrastive loss and the standard cross-entropy loss represents the overall loss minimized during training:

\[
\mathcal{L}_{\text{total}} = \mathcal{L}_{\text{CE}} + \lambda(t) \cdot \mathcal{L}_{\text{con}}
\]

The cross-entropy loss, calculated from the logits $\mathbf{o}_i$, is:

\[
\mathcal{L}_{\text{CE}} = - \frac{1}{|B|} \sum_{i \in B} \sum_{k=1}^K \mathbb{I}[y_i = k] \log \frac{\exp(o_{i,k})}{\sum_{j=1}^K \exp(o_{i,j})}
\]
This combined goal guarantees that the model learns well-structured, separable representations (by $\mathcal{L}_{\text{con}}$) and makes accurate predictions (via $\mathcal{L}_{\text{CE}}$).

\subsubsection{Gradient-Aware Optimization}

While fine-tuning the overall model, we employ alternative learning rates for the newly introduced parameters and the pre-trained RoBERTa parameters. This acknowledges that while new layers can fast learn from scratch, the pre-trained weights already contain important general language knowledge and should be updated carefully. Let $\Theta_{\text{new}}$ represent our new parameters (layer attention $\mathbf{w}$, projection heads, classification head) and $\Theta_{\text{roberta}}$ represent the RoBERTa parameters. The rule for updating is:

\[
\theta \gets \theta - \eta_\theta \cdot \nabla_\theta \mathcal{L}_{\text{total}}, \quad \theta \in \Theta_{\text{roberta}} \cup \Theta_{\text{new}}
\]

We set $\eta_{\text{roberta}} = 2 \times 10^{-6}$ and $\eta_{\text{new}} = 1 \times 10^{-5}$. We use the AdamW optimizer with weight decay.

\paragraph{Full experimental setup}
A RoBERTa-large backbone was used in every experiment. Due to GPU memory limitations, training was done with a batch size of 16 and a maximum input sequence length of 256 tokens. AdamW was used to optimize the model for 15 epochs. Different learning rates were applied to different components to provide stable fine-tuning: $2\times10^{-6}$ for the pre-trained RoBERTa parameters and $1\times10^{-5}$ for newly added layers (layer-attention, projection, and classification heads).  A linear learning rate scheduler with a warm-up phase covering 10\% of the total training steps was used. In the classifier head, dropout regularization was implemented at a rate of 0.4. An adaptive temperature mechanism with base temperature $\tau_{\text{base}} = 0.05$ and scaling factor $\beta = 0.1$ was used for the contrastive learning component. The contrastive loss was included using a progressive weighting schedule, where the weight $\lambda$ rose linearly from 0 to $\lambda_{\max} = 0.15$ for the first $T_{\text{ramp}} = 5$ epochs. Gradient accumulation over two steps and gradient trimming with a maximum norm of 1.0 were used to further stabilize optimization. Following initial tuning, all hyperparameters were chosen based on validation performance.

\subsection{Model Architecture Overview}
\begin{figure*}[h]
    \centering
    \includegraphics[width=0.9\textwidth]{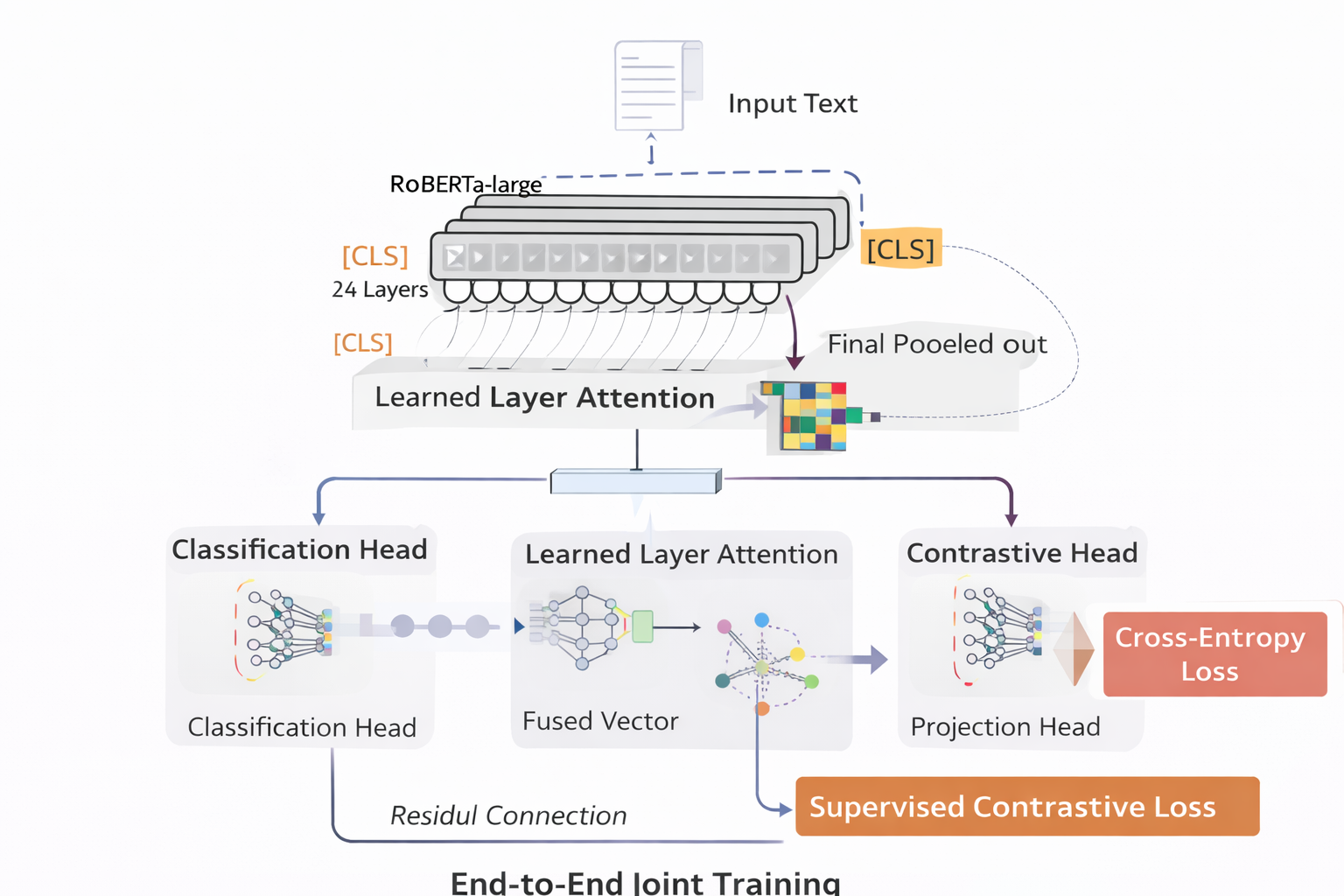} 
    \caption{Complete system architecture.}
    \label{fig:architecture} 
\end{figure*}

The entire data flow of our model is shown in Figure~\ref{fig:architecture}. The RoBERTa-large backbone is traversed by the input text. The final pooled output and the `[CLS]` representations from each of the 24 layers are extracted. A learnt convex combination (layer attention) is used to combine these 25 vectors. After that, this fused representation is employed in two concurrent paths: (1) it is projected onto a contrastive space to calculate the contrastive loss, and (2) it is sent through the classification head to generate logits. The fused representation is stabilized by the remaining link from the last layer. The joint loss is used to train the complete system from beginning to end.

\section{Results and Analysis}

\subsection{Main Results}

We assess our model on the SWMH benchmark against state of the art models. Results for all comparable models are shown in Table~\ref{tab:main_results}. By achieving 72.89\% weighted F1, our suggested model sets a new benchmark.

\begin{table}[htbp]
\centering
\caption{Performance comparison on SWMH test set.}
\label{tab:main_results}
\begin{tabular}{lcc}
\toprule
\textbf{Model} & \textbf{F1 Score (\%)} & \textbf{Recall (\%)} \\
\midrule
BERT & 70.46 & 69.78 \\
RoBERTa & 72.03 & 70.89 \\
BioBERT & 68.60 & 67.10 \\
ClinicalBERT & 68.16 & 67.05 \\
MentalBERT & 71.11 & 69.87 \\
MentalRoBERTa & 72.16 & 70.65 \\
\midrule
\textbf{Our Model} & \textbf{74.36} & \textbf{73.30} \\
\bottomrule
\end{tabular}
\end{table}

Our data shows a significant conclusion. Our model outperforms all baseline models, including domain-specific models that were pre-trained on extensive mental health corpora, such as MentalBERT and MentalRoBERTa. For this task, architectural advances in fine-tuning can surpass the advantages of domain-specific pre-training, as evidenced by the notable 2.20 percentage point improvement over MentalRoBERTa and 3.25 percentage point improvement over MentalBERT. Also notable to mention that our model outperforms it's counterpart finetuned Roberta by 2.33 percentage points.In addition to improving overall F1, the proposed model achieves a 2.41\% improvement in recall over the strongest baseline. This is particularly important in mental health applications, where false negatives may correspond to missed at-risk individuals. The improved recall suggests that the structured embedding space reduces overlap between confusable conditions and enables more consistent detection of minority or subtle cases. To account for randomness in training, each experiment was repeated $n=5$ times using different random seeds. We report the mean performance along with the standard deviation. All results are presented in the form \textit{mean} $\pm$ \textit{standard deviation}.

The proposed model achieved an average F1-score of $74.36\% \pm 0.50$ across all runs.

\subsection{Breaking the Domain Adaptation Barrier}

One particularly noteworthy finding is that our model performs better than MentalRoBERTa, which underwent significant continuous pre-training on 150M tokens from mental health forums~\cite{ji-etal-2022-mentalbert}, utilizing simply standard RoBERTa-large initialization (pre-trained on general online text). This calls into question the widely held belief that optimal performance in specialized domains such as mental health requires domain-adaptive pre-training.

We speculate that this happens because: (1) the contrastive loss reorganizes the representation space to reflect clinical relationships, making up for the absence of domain-specific pre-training by enforcing a semantically meaningful geometry; and (2) our layer-attention mechanism learns to emphasize clinically relevant linguistic features dynamically, effectively performing “soft” domain adaptation during fine-tuning.

\subsection{Component Contributions}

A systematic analysis that isolates the contributions of each suggested component is shown in Table~\ref{tab:ablation} where RoBERTa-large is the base model used in all configurations.

\begin{table}[t]
\centering
\caption{contribution of each component.}
\label{tab:ablation}
\begin{tabular}{lc}
\hline
\textbf{Model Configuration} & \textbf{F1 Score} \\
\hline
Standard Fine-tuning & 72.00 \\
Layer Attention Only & 73.10 \\
Contrastive Loss Only & 73.2 \\
Both (Full model) & \textbf{74.36} \\
\hline
\end{tabular}
\end{table}

Contrastive learning adds 1.10 percentage points, and the layer-attention process adds 1.20 percentage points. Their combination (Full model) achieves F1-score of 74.36\%, slightly exceeding the sum of individual contributions, indicating a modest synergetic effect. and the full model model consistently outperforms all other models in the benchmark as shown in Table~\ref{tab:ablation}.

\section{Conclusion}

In this work, we presented our model, a novel framework for classifying texts related to mental health that combines two complimentary innovations: temperature-scaled supervised contrastive learning and layer-attentive residual connections. By dynamically weighting transformer layers according to their clinical relevance and explicitly structuring the embedding space to reflect clinical relationships between conditions, our method overcomes two basic drawbacks of conventional transformer fine-tuning: inefficient layer utilization and entangled representation spaces.

Our thorough analysis on the SWMH benchmark \cite{ji2021suicidal} shows that our model achieves \textbf{74.36\% weighted F1-score}, setting a new benchmark and outperforming specialized domain-adapted models such as MentalRoBERTa \cite{ji-etal-2022-mentalbert}. This discovery is especially noteworthy since it contradicts the widely held belief that domain-adaptive pre-training is necessary for the best performance in specialized areas \cite{gururangan2020dontstoppretrainingadapt}, implying that advanced fine-tuning techniques can produce better outcomes.

The component contribution show that both elements work together to improve performance. In order to identify small variations between mental health problems, the layer-attention mechanism learns to highlight semantically rich intermediate transformer layers \cite{jawahar-etal-2019-bert}. In the meantime, a representation space where clinically related situations are appropriately clustered and different conditions are well separated is produced by the contrastive loss with adaptive temperature scaling.

our model provides useful benefits for digital mental health applications in addition to its technological contributions. A recurring issue in mental health NLP is addressed by the enhanced separation of confusable condition pairings (such as depression and anxiety), while the interpretable layer attention weights offer transparency into decision-making \cite{shing-etal-2018-expert}.

In order to better reflect clinical reality, future research should consider expanding this framework to multilingual settings \cite{nozza2022challenges}, adding temporal context for longitudinal assessment \cite{DetectingDepression}, and modifying it for spectrum-based rather than categorical classification \cite{FIORILLO2024152502}. Other high-stakes, fine-grained NLP classification jobs might benefit from the architectural advances discussed here.

LARC-RoBERTa is a step toward more dependable and transparent computational tools that can support solving the global mental health crisis by improving the performance and interpretability of mental health text classification. Despite these limitations, our work establishes a strong foundation for representation-aware mental health classification, with future extensions addressing these constraints poised to enhance clinical utility.

\section{Limitations}

There are certain restrictions to our work that should be taken into account. First, our model was trained only on English-language Reddit posts from the SWMH dataset \cite{ji2021suicidal} for the evaluation. Additional validation is needed for generalization to other platforms, languages, and text genres. Second, Our method examines each text post separately, without the chronological context that is frequently essential for evaluating mental health. Comorbidities, longitudinal trends, and symptom progression over time are often taken into account in clinical diagnosis \cite{reece2017detecting}. The existing method may overlook crucial diagnostic information found in consecutive user postings or clinical history since it is unable to capture these dynamic characteristics. 

The datasets used in this research are publicly available through the Hugging face repository.

\section*{Conflict of Interest}

The authors declare that they have no conflict of interest.
\backmatter

\bibliography{sn}

\end{document}